\documentclass[10pt,letterpaper]{article}
\usepackage[top=0.85in,left=2.75in,footskip=0.75in]{geometry}

\usepackage{changepage}

\usepackage[utf8]{inputenc}

\usepackage{textcomp,marvosym}

\usepackage{fixltx2e}

\usepackage{amsmath,amssymb}

\usepackage{cite}

\usepackage{nameref,hyperref}

\usepackage[right]{lineno}

\usepackage{microtype}
\usepackage{rotating}

\raggedright
\usepackage[aboveskip=1pt,labelfont=bf,labelsep=period,justification=raggedright,singlelinecheck=off]{caption}

\usepackage[normalem]{ulem}
\makeatletter
\renewcommand{\@biblabel}[1]{\quad#1.}
\makeatother

\date{}

\usepackage{lastpage,fancyhdr,graphicx}
\usepackage{epstopdf}
\fancyheadoffset[L]{2.25in}
\fancyfootoffset[L]{2.25in}
\begin{document}
\vspace*{0.35in}

% Please capitalize all terms in the title except conjunctions, prepositions, and articles.
\begin{flushleft}
{\Large
\textbf\newline{Modelling Creativity: Identifying Key Components through a Corpus-Based Approach}
}
\newline
\\
Anna Jordanous\textsuperscript{1,\Yinyang, *},
Bill Keller\textsuperscript{2,\Yinyang}
\\
\bigskip
\bf{1} School of Computing, University of Kent, Chatham Maritime, Kent, UK
\\
\bf{2} Department of Informatics, University of Sussex, Falmer, Brighton, UK
\\
\bigskip

% Insert additional author notes using the symbols described below. Insert symbol callouts after author names as necessary.
% 
% Remove or comment out the author notes below if they aren't used.
%
% Primary Equal Contribution Note
\Yinyang These authors contributed equally to this work.

% Additional Equal Contribution Note
% Also use this double-dagger symbol for special authorship notes, such as senior authorship.
%\ddag These authors also contributed equally to this work.

%\textcurrency a Insert current address of first author with an address update
% \textcurrency b Insert current address of second author with an address update
% \textcurrency c Insert current address of third author with an address update


% Group/Consortium Author Note
%\textpilcrow Membership list can be found in the Acknowledgments section.

% Use the asterisk to denote corresponding authorship and provide email address in note below.
* a.k.jordanous@kent.ac.uk / billk@sussex.ac.uk 

\end{flushleft}
\section*{Abstract}

Creativity is a complex, multi-faceted concept encompassing a variety of related aspects, abilities, properties and behaviours. If we wish to study creativity scientifically, then a tractable and well-articulated model of creativity is required. Such a model would be of great value to researchers investigating the nature of creativity and in particular, those concerned with the evaluation of creative practice. This paper describes a unique approach to developing a suitable model of how creative behaviour emerges that is based on the words people use to describe the concept. Using techniques from the field of statistical natural language processing, we identify a collection of fourteen key components of creativity through an analysis of a corpus of academic papers on the topic. Words are identified which appear significantly often in connection with discussions of the concept. Using a measure of lexical similarity to help cluster these words, a number of distinct themes emerge, which collectively contribute to a comprehensive and multi-perspective model of creativity. The components provide an ontology of creativity: a set of building blocks which can be used to model creative practice in a variety of domains. The components have been employed in two case studies to evaluate the creativity of computational systems and have proven useful in articulating achievements of this work and directions for further research.

%\linenumbers

\section*{Introduction}\label{intro}

%Currently, most content on the Semantic Web is in the form of ontologies of `things': semantically structured collections of factual or objective data on topics as diverse as people \cite{foaf}, places \cite{geonames}, music \cite{musicOntology} or manuscripts \cite{sawsOntology}. The focus in Semantic Web research has generally been on reducing ambiguity in representations of these more structured domains and concepts, rather than directly tackling the representation and machine-readable representation of subjective, unstructured concepts in an ontology. However, cognitive science research and work on lexical resources such as WordNet \cite[$http://wordnet.rkbexplorer.com$]{fellbaum98} have laid foundations for more definitionally troublesome concepts to be considered in detail. The time is ripe for developing ontologies of subjective concepts such as creativity.

What is creativity, and how can we better understand and learn about creativity using computational modelling? Computational creativity is a relatively youthful research area that has been growing with significant pace in recent years. Computational creativity is: 
\begin{quote}`The philosophy, science and engineering of computational systems which, by taking on particular responsibilities, exhibit behaviours that unbiased observers would deem to be creative.' 
\cite[p. 21]{coltonwiggins12}
\end{quote}

Computational creativity research follows both theoretical and practical directions and crosses several disciplinary boundaries between the arts, sciences, and engineering. Research within the field is influenced by artificial intelligence, computer science, psychology and specific creative domains that have received attention from computational creativity researchers to date, such as art, music, reasoning and narrative/story telling \cite[provide examples]{colton08,widmerAIM09,leon10MM,perezyperez99}. 

The evaluation of creative systems developed by researchers in the field of computational creativity has proven non-trivial. Creativity evaluation, a recurring topic for discussion, has been described as a `Grand Challenge' for computational creativity research \cite{cardoso09}. Difficulties are inherently linked to a question that both motivates and complicates the computational modelling of creativity: what  do we mean when we talk about `creativity' and what does it constitute?

Creativity is a complex, multi-faceted concept encompassing a variety of related aspects, abilities, properties and behaviours. There have been many attempts to capture this concept in words; indeed the work described in this paper is based on thirty such attempts (see the Methods section and the papers listed in \nameref{S1_Appendix}). 
In the academic literature on creativity, many common themes have emerged. 
However, multiple viewpoints exist, prioritising different aspects of the concept according to what are traditionally considered to be the primary factors for a particular discipline. 
The need for a more over-arching, inclusive, multi-dimensional account of creativity has been widely recognised \cite{rhodes61,torrance67,plucker04defn,kaufman09}. Such a meta-level account would assist our understanding of creativity, highlighting areas of common ground and avoiding the pitfalls of disciplinary bias \cite{hennessey10,plucker04}.

There are many challenges to modelling a concept like creativity in a computational setting. Conceptually, creativity seems inherently fuzzy or vague, with a meaning that shifts depending on the domain of application. Tackling these challenges affords two key advantages, both of which motivate the current paper. First, we can take advantage of computing and artificial intelligence to perform or enhance creative activities using computational power and research expertise. Secondly, the act of modelling creativity requires us to more carefully identify what informs our intuitive notions about creativity and this can guide us towards a more rigorous and comprehensive understanding of the concept.  

The aim of the work reported in this paper is to examine the nature of creativity and to identify within it a set of components, representing key dimensions, that are recognised across a combination of different viewpoints. We present a novel, empirical approach to the problem of modelling how creative behaviour is manifested, that focuses on what is revealed about our understanding of creativity and its attributes by the words we use to discuss and debate the nature of the concept. Analysis of this language provides a sound basis for constructing a sufficiently detailed and comprehensive model of creativity \cite{lakoff87,wittgenstein58}. The current work is intended as a significant, methodological contribution towards addressing the Grand Challenge of evaluation in computational creativity research. It should provide researchers with a firm foundation for evaluating exactly how creative so-called {\emph creative systems} actually are.

On our approach, statistical language processing techniques are used to identify words significantly associated with creativity in a corpus of academic papers on the subject. A corpus spanning some 60 years of research into the nature of creativity was collected together. The papers were gathered from a wide variety of disciplines including psychology, educational testing and computational creativity, amongst others. The language data drawn from this collection was then analysed and contrasted with data from a corpus of matched papers on subjects unrelated to creativity. From this analysis, a set of 694 {\em creativity words\/} was identified, where each creativity word appeared significantly more often than expected in the corpus of creativity papers. A measure of lexical similarity provided a basis for clustering the creativity words into groups of words with similar or shared aspects of meaning. Through inspection of these clusters, a total of fourteen {\em key components of creativity\/} was identified, where each represents a key theme or attribute of creativity. The set of components yields information about the nature of creativity, based on what is collectively emphasised in discussions about the concept. 

% The Semantic Web has emerged as a way to address the troublesome but important issue \cite{boden99} of articulating values, concepts and information using an open and  {\em machine-understandable} vocabulary. Encoding the creativity components in an OWL ontology has enabled this {\em ontology of creativity} to be made publicly available to the wider research community as a resource in the Semantic Web, under the permanent URL:
%\begin{quote}
%{\small \texttt{http://purl.org/creativity/ontology}}
%\end{quote}

In the rest of this section we begin by noting a variety of attempts to define creativity. The representation of subjective, ambiguous, loosely structured concepts is considered.
%, both from the perspective of current Semantic Web research and from a broader conceptual perspective. 
In the remaining sections, details are provided of the methodology used to identify components of creativity from an analysis of language data. The results of this analysis are then presented in terms of a model that encompasses fourteen key components. The derived set of components is evaluated in terms of how well it satisfies the need for a shared, inclusive and comprehensive account of creativity and provides a vocabulary of creativity that is accessible to both people and machines. Finally, conclusions are drawn and some directions for further work are outlined.

\subsection*{Background:  The nature of creativity} \label{background}

As Torrance observes:

\begin{quote}
`[c]reativity defies precise definition ... even if we had a precise conception of creativity, I am certain we would have difficulty putting it into words' \cite[p. 43]{torrance88}. 
\end{quote}

Many other authors have expressed similar difficulties \cite{rhodes61,sternberg99a,kaufman09}. 
In their review of research into human creativity, Hennessey and Amabile ask a significant follow-on question: 
\begin{quote}
`Even if this mysterious phenomenon can be isolated, quantified, and dissected, why bother? Wouldn't it make more sense to revel in the mystery and wonder of it all?' 
\cite[p. 570]{hennessey10}
\end{quote}

\noindent Two answers to this question are offered by Hennessey and Amabile, both of which are identified as desirable: to gain a deeper understanding of creativity and to learn how to boost people's creativity.

Creativity can and should be studied and measured scientifically, but the lack of a commonly-agreed understanding causes problems for measurement \cite{kaufman09}. Plucker et al. make recommendations about best practice based on their own survey of the creativity literature:
\begin{quote}
\noindent
`we argue that creativity researchers must

\begin{enumerate}
\renewcommand{\labelenumi}{(\alph{enumi})}
\item explicitly define what they mean by creativity,
\item avoid using scores of creativity measures as the sole definition of creativity (e.g., creativity is what creativity tests measure and creativity tests measure creativity, therefore we will use a score on a creativity test as our outcome variable),
\item discuss how the definition they are using is similar to or different from other definitions, and
\item address the question of creativity for whom and in what context.' \cite[p.92]{plucker04defn}
\end{enumerate}
\end{quote}

\noindent In short, we need to specify and justify the standards that we use to judge creativity. A more objective and well-articulated account of how creativity is manifested enables researchers to make a worthwhile contribution \cite{torrance67,plucker04defn,kaufman09}. Particularly, in research we would like to focus on what processes and concepts relevant to creativity are `sufficiently important to warrant study' \cite[p. 15]{vartanian14}, based on an {\em accumulation} of the body of work on creativity to date \cite{vartanian14}.

\paragraph{Definitions of creativity.}
\label{existing_defs}

To find out the meaning of a word, a natural first step might be to consult a dictionary. Dictionary definitions of creativity provide a brief introduction to  the meaning of the word. However, for the purposes of research, the utility of such definitions is severely restricted by their format and brevity, and they generally provide only cursory, shallow insights into the nature of creativity. More problematic still, dictionary entries are often self-referential or circular,  defining creativity in terms of ``being creative'' or ``creative ability''. To illustrate these limitations, there follow several typical dictionary definitions of creativity and the related words creative and create:\footnote{For readability, some definitions are edited slightly to standardise formats and remove etymological/grammatical annotations.}
\begin{quotation}
\begin{description}
\item {\em Oxford English Dictionary} 2nd ed. (1989) pp. 1134-5:
	\begin{enumerate}
	\item[creativity:] creative power or faculty; ability to create
	\item [creative:] Having the quality of creating, able to create; of or relating to creation; originative. b. Inventive, imaginative; of, relating to, displaying, using, or involving imagination or original ideas as well as routine skill or intellect, esp. in literature or art. c. Esp. of a financial or other strategy: ingenious, esp. in a misleading way. 2. Providing the cause or occasion of, productive of.
\item [create:] 1.a. Said of the divine agent: To bring into being, cause to exist; esp. to produce where nothing was before, 'to form out of nothing'. b. with complemental extension. 2. To make, form, constitute, or bring into legal existence (an institution, condition, action, mental product, or form, not existing before). Sometimes of material works. 3. To constitute (a personage of rank or dignity); to invest with rank, title, etc. 4. To cause, occasion, produce, give rise to (a condition or set of circumstances).
	\end{enumerate}
\item {\em The Penguin English Dictionary} 2nd ed. (1969) p. 174:
	\begin{enumerate}
	\item[creativity:] creative power or faculty; ability to create
	\item[creative:] having power to create; related to process of creation; constructive, original, producing an essentially new product; produced by original intellectual or artistic effort 
	\item [create:] make out of nothing, bestow existence on; cause, bring about; produce or make something new or original; confer new rank etc on; (theat) be the first to act (a certain part); make a fuss
	\end{enumerate}
\item {\em Webster's 3rd New International Dictionary} (1961) p. 532:
	\begin{enumerate}
	\item[creativity:] the quality of being creative; ability to create 
	\item[creative:] 1. having the power or quality of creating; given to creation 2: PRODUCTIVE - used with 3: having the quality of something created rather than imitated or assembled; expressive of the maker; IMAGINATIVE 
	\item[create:] 1: to bring into existence; make out of nothing and for the first time 2: to cause to be or to produce by fiat or by mental, moral, or legal action 3: to cause or occasion - used of natural or physical causes and esp. of social and evolutionary or emergent forces 4a: to produce (as a work of art or of dramatic interpretation) along new or unconventional lines) b: to design (as a costume or dress)
	\end{enumerate}
\end{description}
\end{quotation}

Given the problems inherent in dictionary definitions of creativity, it is not surprising that a number of creativity researchers have set out  to provide their own definitions of the concept.  Some examples are:

\begin{quotation}
\begin{quote}
 `creativity is that process which results in a novel work that is accepted as tenable or useful or satisfying by a group at some point in time' %By this definition we limit ourselves to the study of individuals who are regarded as creative by ``significant others'' in their environment.'
\cite[p. 218]{stein63}
\end{quote}
%\begin{quote}
%`A product or response will be judged as creative to the extent that (a) it is both a novel and appropriate, useful, correct or valuable response to the task at hand, and (b) the task is heuristic rather than algorithmic.'
%\cite[p. 35]{amabile96}
%\end{quote}
\begin{quote}
`Creativity is the ability to produce work that is both novel (i.e., original, unexpected) and appropriate (i.e., useful, adaptive concerning task constraints)'
\cite[p. 3]{sternberg99a}
\end{quote}
\begin{quote}
`Creativity is the ability to come up with ideas or artefacts that are {\em new}, {\em surprising and valuable}'
\cite[p. 1]{boden04}
\end{quote}
\begin{quote}
`Creativity is the interaction among \emph{aptitude, process, and environment} by which an individual or group produces a \emph{perceptible product} that is both \emph{novel and useful} as defined within a \emph{social context}'
\cite[p. 90]{plucker04defn}
\end{quote}
\begin{quote}
`Creativity: the generation of products or ideas that are both novel and appropriate' 
\cite[p. 570]{hennessey10}
\end{quote}
\begin{quote}
`The word creativity is a noun naming the phenomenon in which a person communicates a new concept (which is the product). Mental activity (or mental process) is implicit in this definition, and of course no one could conceive of a person living or operating in a vacuum, so the term \emph{press} is also implicit'
\cite[p. 305]{rhodes61}
\end{quote}
\end{quotation}

These more research-oriented definitions avoid the problems of self-reference and circularity noted for the dictionary entries given previously. However, whilst the definitions may provide somewhat deeper insight into the nature of creativity, the brevity of the definitions means that they still only succeed in providing shallow, summary accounts of the concept. 

\paragraph{A multitude of different perspectives} \label{perspectives}

The problem of identifying and quantifying creativity exists across many disciplines. How creative is this person? Does this person have the creative abilities to boost my business? Is this pupil's story creative? Is this computational system an example of computational creativity? As a consequence, when attempts are made to define creativity, it is often from the perspective of a particular domain or research discipline. For example, psychometric tests for creativity such as \cite{guilford50,torrance74} focus on {\em problem solving} and {\em divergent thinking} as key attributes of a creative person. In contrast, computational creativity research \cite[for example]{pease01,wiggins06ngc,peinado06ngc,ritchie07} has historially placed emphasis on the {\em novelty} and {\em value} of creative products. Whilst there is some consensus across academic fields, for example novelty and value are typically recognised as necessary (but arguably not sufficient) components of creativity \cite{mayer99}, the differing emphases contribute to variations in the interpretation of creativity. These variations affect consistency across creativity research in different disciplines and potentially hinder interdisciplinary collaborations and cross-application of findings.

Several competing interpretations of creativity exist in the literature. Sometimes these differences of opinion do not need to be directly resolved but can be included alongside each other. Examples include whether creativity is centred around mental processes \cite{boden04,dietrich10,gabora12} or embodied and situated in an interactive environment \cite{mccormack07,sosa09}. Another example is whether creativity is domain-independent \cite{plucker98}, or dependent on domain-specific context \cite{baer98}, or (as both Plucker and Baer have concluded) a combination of both \cite{plucker04,baer10}. 

Other conflicts arise where a previously narrow view of creativity has been widened in perspective. To resolve the conflict, an inclusive, all-encompassing view of creativity should adopt the wider perspective and incorporate the narrower perspective. For example rather than focussing narrowly on creative {\em genius}, through the study of people with exceptional creative achievements \cite[for example]{poincare29,hadamard45} emphasis has shifted to encompass the broader study of {\em everyday} creativity, with genius as a special case: the notion that everyone can be creative to some degree \cite{weisberg88,bryankinns09}.

Similarly, researchers distinguish between {\em little-c} and {\em Big-C} creativity, or {\em psychological/P-creativity} and  {\em historical/H-creativity} \cite{boden04}, adjusting their focus accordingly to make their research more manageable. This is particularly the case in computational creativity, where endowing the computer with elements of general, human knowledge and experience is a major challenge. Little-c creative or p-creative work is perceived as creative by the creator personally but may replicate existing work (unknown to the creator) so is not necessarily creative in a wider social context. This encompasses the concept of Big-C creativity or h-creativity, where the work makes a creative contribution both to the creator and to society. To be Big-C creative/h-creative is to be little-c creative/p-creative in a way which has not been done before by anyone.

The preceeding discussion indicates that creativity is a complex, multi-faceted concept that requires a broad and inclusive treatment. The {\em Four Ps} framework \cite{rhodes61,stein63,mooney63,odena09,jordanous16cs} ensures we pay attention to four key aspects of creativity:

\begin{description}
\setlength{\labelwidth}{1.5cm}
\setlength{\itemindent}{1.5cm}
\item[Person/Producer:] The individual that is creative
\item[Process:] What the creative individual does to be creative
\item[Product:] What is produced as a result of the creative process
\item[Press:] The environment in which creative activity takes place
\end{description}

\noindent This framework presents creativity in a broader context, making our understanding of the concept more generally applicable and less specific to a domain or academic discipline. In contrast, models of the creative process \cite{wallas26,poincare29,hadamard45}, tests of people's creativity \cite{goldman64,guilford67,torrance74} or tests based on creative artefact generation \cite{amabile96,ritchie07} are useful only within a limited sphere. Jordanous \cite{jordanous16cs} has contextualised the Four Ps in a computational context, referring to the creative Producer (person or computational agent) carrying out Processes within the environmental context of a Press, to create computational Products.

\subsection*{The challenges of modelling creativity} \label{cogPhil}

% Was Reflections on the semantics of subjective concepts. Merge this together and make it more relevant to modelling creativity (specifically) rather than the general treatment it is now.

Creativity can be seen as an essentially contested concept \cite{gallie56}: it is subjective, abstract and can be interpreted in a variety of acceptable ways, such that a fixed `proper general use' is elusive \cite[p. 167]{gallie56}. Gallie \cite{gallie56} defines an essentially contested concept through several features: being internally complex in nature, but amenable to being broken down into identifiable constituent elements of varying relative importance, and dependent on a number of factors such as context and individual preference. Although there may be consensus on the meaning of such concepts in very general terms, they may defy precise interpretation. There is not a single agreed instantiation, but instead many reasonable possibilities, influenced by changing circumstances and contexts. It is more productive to acknowledge that these different interpretations exist and refer to `the respective contributions of its various parts or features' \cite[p. 172]{gallie56}, rather than to argue for a single interpretation. Thus, different types of creativity manifest themselves in different ways while sharing certain characteristics (not necessarily the same across all creative instances). This is what Wittgenstein refers to as `family resemblances' \cite{wittgenstein58}:

\begin{quote}
[On discussing the example of what a `game' is] `we see a complicated network of similarities overlapping and criss-crossing: sometimes overall similarities, sometimes similarities of detail. ... I can think of no better expression to characterize these similarities than ``family resemblances''; for the various resemblances between members of a family: build, features, colour of eyes, gait, temperament, etc. etc. overlap and criss-cross in the same way. And I shall say: ``games'' form a family.'
\cite[Part 1, Paragraphs 66-67]{wittgenstein58}
\end{quote}

Similarly, with creativity, different manifestations of creativity are not all necessarily required to share the same common, core elements in order to be identified as part of the creativity `family'. Rather, relationships between different manifestations reveal various shared characteristics that emerge in a similar way to Wittgenstein's `family resemblances' in language. We need to identify what those family resemblances are in the case of creativity. To understand creativity, we can investigate what resemblances exist across different instantiations of the concept. 

Wittgenstein \cite{wittgenstein58} has argued that `a clear view of the aim and functioning of the words' helps us `dispers[e] the fog' that obscures a clear vision of the `working of language' \cite[Part 1, Paragraph 5]{wittgenstein58}. To understand the use of a word, one must have background information and context. Wittgenstein gives the example of a chess piece, which is introduced to someone as a `king' \cite[paragraph 31]{wittgenstein58}: to understand this usage the person must already know the rules of chess, or must at least know what it means to have a piece in a game. To Wittgenstein, the semantics of words and statements are determined by how we use them, grounded in rules set by our habitual use of a word and our shared consensual practices, rather than being fixed by static, pre-assigned meanings.        

%Waismann, a contemporary of Wittgenstein, has reflected on the impact of `open texture' on language, where certain words or phrases simply cannot be completely defined for all possible scenarios:

%\begin{quote}
%`Every description stretches, as it were, into a horizon of open possibilities: however far I go, I shall always carry this horizon with me.'
%\cite[p. 122]{waismann68}
%\end{quote}
Linguistics research advocates that the meaning of a word is dependent on the context it is used in \cite{firth57}. In particular, Lakoff has argued that the study of language helps reveal how people think \cite{lakoff87,lakoff80}. Words used frequently in discussions of the nature of a concept provide the context for the commonly understood meaning of that concept, as has been shown in various corpus linguistics contributions \cite{oakes98,rayson00,kilgarriff01,kilgarriff06}. 

%We acknowledge that semantic ambiguity is unlikely to be resolved to a single agreed definition. Meanings can be dynamic, evolving over time and contextually dependent, rather than being fixed, with a static and unchanging prescribed definition. 
The key principle emerging across these present discussions is that the meaning of words like creativity can be modelled by identifying different aspects that collectively contribute to the meaning of the concept of creativity.

The need for a clearer, multi-perspective understanding of creativity is evident, but remains to be addressed. There is a large quantity of material contributory to a satisfactory model of creativity and a number of key contributions have been discussed during this section. What must be done now is to marshal this assortment of material and to unify different perspectives where possible, in order to avoid the disciplinary `blinkers' or compartmentalisation that is often seen in creativity research \cite{hennessey10}. In approaching the semantic representation of subjective and multi-faceted concepts, some useful guidance is offered through philosophical reflections on the meaning of such concepts.

%Further, cognitive science methodologies, {\bf IS THIS NEXT BIT TO GO?} as well as approaches to ontology learning from texts \cite{gomezperez04} {\bf END} can be harnessed to enable the representation of more semantically ambiguous concepts such as creativity

%To date, however, the Semantic Web research community has largely focused on the representation of objective concepts and resources that are relatively clearly defined. 

% You may title this section "Methods" or "Models". 
% "Models" is not a valid title for PLoS ONE authors. However, PLoS ONE
% authors may use "Analysis" 
\section*{Methods}
\label{methodology}

Our approach makes use of an empirical study and analysis of the language used to talk about creativity in order to gather and collate knowledge about the concept. In addition, following from the observations above, a {\em  confluence approach\/} to creativity is adopted \cite{sternberg99a,mayer99,ivcevic09}. This works on the principle that creativity results from several components converging and goes on to examine what these components are. Taking this approach in conjunction with the application of tools from computational linguistics and statistical analysis allows a wider disciplinary spectrum of perspectives on creativity to be captured than has previously been attempted. This is achieved by breaking down the whole into smaller and more tractable constituent parts identified through a broad cross-disciplinary examination of creativity research.

Tools from natural language processing and statistical analysis are used to identify words that appear to be highly associated with dimensions of creativity, as represented in a sample of academic papers on the topic. A key innovation is the use of a statistical measure of lexical similarity, which allows the words to be clustered into coherent and semantically-related groups. Clustering reveals a number of common themes or factors of creativity, allowing the identification of a set of fourteen components that serve as building blocks for creativity.

\subsection*{Corpus data}\label{corpusData}

A sample of academic papers discussing the nature of creativity was assembled as a {\em creativity corpus\/} in 2010. This creativity corpus consisted of 30 papers examining creativity from various academic stand-points ranging from psychological studies to computational models. 

\vspace{0.25cm}
\begin{center}
\fbox{
\begin{minipage}{0.8\linewidth}
\textbf{Creativity corpus}: \\ a collection of thirty academic papers which explicitly discuss the nature of creativity.
\end{minipage}
}
\end{center}
\vspace{0.25cm}

The 30 papers selected for the creativity corpus are listed in \nameref{S1_Appendix}. The strategy used to select papers for this corpus is illustrated in a flow diagram, in Fig. \ref{searchCC}. 

\begin{figure}[!b]
\caption{{\bf A flow diagram describing the search strategy used to identify papers for the creativity corpus.}}
% REMOVE IMAGES FROM SUBMITTED VERSION
%\includegraphics[width=\textwidth]{images/Fig1}
\label{searchCC}
\end{figure} 

The search strategy for identifying papers for the corpus involved a literature search for the term `creativity' on the academic database {\em Scopus} to identify suitable papers. This literature search was supplemented with additional influential papers which may not have appeared in a Scopus search; for example, a computer science conference paper on cognitive models of creativity.\footnote{It should be noted that in Computer Science, a number of conferences carry as much or more publication weight as some journals in the field.} The eligibility of each identified article was verified for inclusion in the corpus via careful manual inspection.

Paper selection for the creativity corpus was governed by inclusion criteria based on measuring the influence of a paper and coverage of a wide range of years and academic disciplines. The inclusion criteria are as follows, listed in descending order of precedence: 
\begin{itemize}
\item Papers must have, as their primary focus, discussion of the nature of creativity.
\item Papers should be considered particularly influential. Influence was generally measured objectively, in terms of the number of times a paper had been cited by other academic authors. However, for papers published in recent years and which had consequently had little time to accrue citations, selection was based instead on a subjective judgement of influence grounded in a knowledge of the field.
\item Papers selected should, as far as reasonable, represent a cross-section of years over the range 1950-2009. [The corpus was compiled in 2010.] 1950 was chosen as a starting point in recognition of the effect of J. P. Guilford's presidential address to the American Psychological Association \cite{guilford50}, which examined contemporary creativity research (or more specifically, the lack of thereof). His talk was highly influential in encouraging more creativity research activity \cite{kaufman09}. 
\item Papers selected should, as far as reasonable, represent a cross-section of disciplines relevant to discussions of creativity. Fig. \ref{disciplinaryDistribution} illustrates the disciplinary distribution of the corpus as it changes over the time period covered by the selected papers. This distribution is based on the {\em Scopus} database, which classifies journals under their main subject area(s) covered. We should acknowledge here though that while many disciplines include creative practice, often the focus is on application rather than in depth discussion of what creativity entails. Hence, while we sought to cover creativity from a broad range of perspectives, we also felt it was important not to compromise the focus of our corpus as a representation of key discussions about the nature of creativity.
\end{itemize}

\begin{figure}[!b]
% REMOVE IMAGES FROM SUBMITTED VERSION
%%\includegraphics[scale=0.5]{images/Fig2}
\caption{{\bf Representation of the disciplinary breakdown of the Creativity Corpus by time period.} \\Disciplines are as specified for the paper's journal, by the academic database {\em Scopus}. Note that Scopus may classify a journal under more than one discipline.}
\label{disciplinaryDistribution}
\end{figure}

Exclusion criteria for this search were as follows: 
\begin{itemize}
\item Authors were only represented more than once in the corpus if the relevant papers were written from different perspectives. For example, Mark Runco's work is represented twice in the corpus, but covering two different topics relating to the nature of creativity (psychoeconomic approach to creativity; cognition and creativity). If the search process highlighted two or more papers with a shared author on the same or highly similar perspectives on creativity, then the more highly cited paper was chosen. 
\item Papers had to be written in English, as the language processing tools we were working with were for English language texts.
\item Papers had to be available in a format that enabled us easily to extract plain text (this excluded books or book chapters).
\end{itemize}

The creativity corpus is relatively small and necessarily selective  in terms of the papers that are included. As such it constitutes just a small fraction of the many academic works on creativity that have been published in the last 60 or so years. Indeed, the 30 papers in the creativity corpus cannot be regarded as comprehensively representative of the wide range of academic positions on creativity that have been discussed in the literature over the decades. However, the goal of this work is not to present a fine-grained analysis of language use drawn from this complete literature, nor to provide a comprehensive lexicon or dictionary of creativity. Rather, the goal is to identify the broader ontological themes or factors that recur in our understanding of the concept of creativity. For this purpose, what is required is a sufficiently representative sample of the academic discourse on creativity. This sample can be used to identify the way in which word use reflects key themes or factors that persist across different perspectives.

Our objective is to identify what is distinctive in the language used to discuss creativity, in contrast to the language used to discuss other topics. As a basis for comparison, therefore, a further sample of 60 academic papers on topics unrelated to creativity -- the {\em non-creativity corpus} -- was assembled alongside the creativity corpus, in 2010. 

\vspace{0.25cm}
\begin{center}
\fbox{
\begin{minipage}{0.8\linewidth}
\textbf{Non-creativity corpus}: \\ 60 academic papers on topics unrelated to creativity, from the  same range of academic disciplines and publication years as  the creativity corpus papers.
\end{minipage}
}
\end{center}
\vspace{0.25cm}

The non-creativity corpus papers were selected by a literature search retrieving, for each paper in the creativity corpus, the two most-cited papers in the same academic discipline (as categorised by  {\em Scopus}) and published in the same year, that did not contain any words with the prefix {\em creat\/} (i.e. {\em creativity\/}, {\em creative\/}, {\em creation\/}, and so on). In other words, the criteria for inclusion in this second corpus were whether a paper was one of the two papers that was most highly cited at the time of the search (2010), in the same academic discipline, and published in the same year, as a paper in the creativity corpus, and that satisfied the exclusion criteria of not containing any words with the above mentioned prefixes. The 60 papers selected for the creativity corpus are listed in \nameref{S2_Appendix}. The search strategy used to select papers for this non-creativity corpus is illustrated in a second flow diagram, in Fig. \ref{searchNonCC}. 

\begin{figure}
\caption{ A flow diagram describing the search strategy used to identify papers for the non-creativity corpus.}
% REMOVE IMAGES FROM SUBMITTED VERSION
%\includegraphics[width=\textwidth]{images/Fig3}
\label{searchNonCC}
\end{figure}

The non-creativity corpus is twice the size of the creativity corpus ($\approx$ 700,000 words and $\approx$ 300,000 words respectively), in acknowledgement of the fact that in general the set of academic papers on creativity is only a small subset of all academic papers. Both corpora are very small in comparison to corpora such as the British National Corpus, a relatively large  ($\approx$ 100M words) corpus of written and spoken English in general usage across a number of different contexts, and tiny in comparison to more recent web-derived text collections containing billions of words. There are, however, several benefits associated with using a corpus derived from specialist academic literature:

\begin{itemize}
\item Ease of locating relevant and appropriate papers: e.g. availability of tools to perform targeted literature searches, electronic publication of papers for download, tagging of paper content by keywords, citations in papers to other related papers.
\item Ability to access timestamped textual materials over a range of decades.
\item Publication of academic papers in an appropriate format for computational analysis: most papers that are available electronically are in formats such as PDF or HTML, which can be converted to text fairly easily.
\item Availability of citation data as a measure of how influential a paper is on others: whilst not a perfect reflection of a paper's influence, citation data is often used for measuring the impact of a journal \cite{garfield72} or an individual researcher's output \cite{hirsch05}.
\item Availability of provenance data, such as who wrote the paper and for what audience (from the disciplinary classification of the journal).
\end{itemize}
\noindent 

Some pre-processing was undertaken for each paper in both the creativity corpus and non-creativity corpus prior to analysis. A plain text file was generated for each paper, containing the full text of that paper. All journal headers and copyright notices were removed from each paper, as were the author names and affiliations, list of references and acknowledgements. All files were also checked for any non-ASCII characters and anomalies that may have arisen during the creation of the text file.

\subsection*{Natural language processing}

The corpus data was first pre-processed using the RASP natural language processing toolkit \cite{briscoe06} in order to perform {\em lemmatisation} and {\em part-of-speech} tagging.  Lemmatisation permits inflectional variants of a given word to be identified with a common `dictonary headword' form or `lemma'.  For example,  {\em performs\/}, {\em performed\/}  and {\em performing\/} all occur in the creativity corpus as distinct morphological variants of  the  verb, {\em perform\/}.  Intuitively, we would like to count each of these inflectional variants as an instance of the same word, rather than as separate and distinct lexical tokens. Lemmatisation software enables us to do this by mapping such variants to a cannonical lemma form. As a further refinement, each lemma was also mapped to lower case to ensure that capitalised word forms (e.g. {\em Novel\/}) were not counted separately from their non-capitalised forms ({\em novel\/}).\footnote{While this can result in confusion between proper names and common nouns (e.g. {\em Apple\/} v. {\em apple\/}), it is not considered that the resulting level of 'noise'  in the data is likely to adversely affect the results of the analysis.}

Each word was assigned a part-of-speech tag identifying its grammatical category (i.e. whether the word was a noun, verb, preposition, etc.). Such tagging is useful because it allows us to distinguish between different grammatical uses of a common orthographic form. For example, the use of {\em novel\/} as a noun in {\em a good novel\/} can be properly differentiated from its use as an adjective in {\em a novel idea\/}.  The data was further simplified and filtered so that only words of the four `major' categories (i.e. noun, verb, adjective and adverb) were represented. Note that the major categories bear the semantic content of the papers making up the creativity corpus. They may be distinguished from minor categories or `function words', such as pronouns ({\em something\/}, {\em itself\/}) prepositions (e.g. {\em upon\/}, {\em by\/}) conjunctions ({\em but\/}, {\em or\/}) and quantifiers (e.g. {\em many\/}, {\em more\/}). Because such words have little independent semantic content, they are of limited interest for the present study and may be removed from the data.  

Following processing with RASP, a list of words found in the creativity corpus, together with their frequency counts was generated. The non-creativity corpus was pre-processed in the same way and a corresponding list of words and frequencies also generated.

\subsection*{Identifying words associated with creativity}

The word frequency data derived from the two corpora was used to establish which words occur significantly more often in the creativity corpus than in the non-creativity corpus. This in turn can be regarded as providing evidence of which words are salient to the concept of creativity. Salient words were identified using the log-likelihood ratio  (also referred to as the $G^2$ or G-squared statistic), which is a measure of how well observed data fit a model or expected distribution \cite{dunning93,kilgarriff01,rayson00,oakes98}. It provides an alternative to Pearson's chi-squared ($\chi^2$) test and has been advocated as the more appropriate measure of the two for corpus analysis as it does not rely on the (unjustifiable) assumption of normality in word distribution  \cite{dunning93,kilgarriff01,oakes98}. This is a particular issue when analysing smaller corpora, such as those used in the present work. 
The log likelihood ratio statistic is more accurate in its treatment of infrequent words in the data,  which often hold useful information. By contrast, the $\chi^2$ statistic tends to under-emphasise such outliers at the expense of very frequently occurring data points.

Our use of the log-likelihood ratio follows that of Rayson and Garside \cite{rayson00}.
Given two corpora (in our case, the creativity corpus $cc$ and the non-creativity corpus $nc$) the log-likelihood score for a given word is calculated as shown in equation~(\ref{LLR}) below:

\begin{equation}
\mbox{\em LL} =  O_{cc}\ln (\frac{O_{cc}}{E_{cc}} ) + O_{nc}\ln (\frac{O_{nc}}{E_{nc}} )
\label{LLR} 
\end{equation}
\noindent
where $O_{cc}$ ($O_{nc}$) is the observed frequence of the word in $cc$ ($nc$) and similarly $E_{cc}$ ($E_{nc}$) is its expected frequency. The expected frequency $E_{cc}$  is given by: 

\begin{equation}
E_{cc} = \frac{N_{cc} \times (O_{cc} + O_{nc})}{N_{cc} + N_{nc}}
\label{modelFreqs} 
\end{equation}
where $N_{cc}$ denotes the total number of words in corpus $cc$ (i.e. the sum of the frequencies of all words drawn from corpus $cc$). The expected frequency $E_{nc}$ is defined in a way analogous to Equation \ref{modelFreqs}.

As computed above, the log-likelihood ratio measures the extent to which the distribution of a given word deviates from what might be expected if its distribution is not corpus dependent. The higher the log likelihood ratio score for a given word, the greater the deviation from what is expected. It should be noted however, that the statistic tells us only that  the observed distribution of a word in the two corpora is unexpected (and to what extent).  It does not  tell us whether the word is more or less frequent than expected in the creativity corpus. To identify words significantly associated with creativity therefore, it was necessary to select just those words with observed counts higher than that expected in the creativity corpus. It should perhaps be further noted that the resulting words may be either positively or negatively connoted with respect to creativity. In practice this is not a problem, as the significance of a given word lies in its semantic connection to creativity, not in its sentiment or affect. Affect is taken into account as part of the later manual examination of the data used to identify components of creativity.

The results of the calculations were filtered to remove any words with a log-likelihood score less than 10.83, representing a chi-squared significance value for p=0.001 (one degree of freedom). In this way, the filtering process reduced the set of candidate words to just those that appear to occur significantly more often than expected in the creativity corpus. To avoid extremely infrequent words disproportionately affecting the data, any word occurring fewer than five times was also removed from the data. Finally, the words were inspected to remove any `spurious' items such as proper nouns or misclassified or odd character sequences. This resulted in a total of 694 {\em creativity words\/}: a collection of 389 nouns, 205 adjectives, 72 verbs and 28 adverbs that occurred significantly more often than expected in the creativity corpus. Table \ref{creativityTop20} gives the top 20 results of these calculations.

\begin{table}[!b]
\begin{center}
\begin{tabular}{c c | c c}
\# & Word (\& POS tag) & LLR \\
\hline
\#1 & thinking (N) & 834.55  \\
\#2 & process (N) & 612.05 \\
\#3 & innovation (N) & 546.20 \\
\#4 & idea (N) & 475.74  \\
\#5 & program (N) & 474.41  \\
\#6 & domain (N) & 436.58 \\
\#7 & cognitive (J) & 393.79  \\
\#8 & divergent (J) & 355.11 \\
\#9 & openness (N) & 328.57  \\
\#10 & discovery (N) & 327.38 \\
\#11 & primary (J) & 326.65 \\
\#12 & originality (N) & 315.60  \\
\#13 & criterion (N) & 312.61  \\
\#14 & intelligence (N) & 309.31 \\
\#15 & ability (N) & 299.27 \\
\#16 & knowledge (N) & 290.48  \\
\#17 & create (V) & 280.06 \\
\#18 & experiment (N) & 253.32 \\
\#19 & plan (N) & 246.29 \\
\#20 & agent (N) & 246.24  \\
\end{tabular}
\end{center}
\caption{{\bf The top 20 results (in descending order) of the log-likelihood ratio (LLR) calculations.}\\ A significant LLR score at p=0.001 is 10.83. N.B. POS=Part Of Speech: N=noun, J=adjective, V=verb, R=adverb.}
\label{creativityTop20}
\end{table}%

\subsection*{Identifying components of creativity}
\label{creat-comps}

It is important to note that our objective is to identify key themes in the lexical data, not to induce a comprehensive terminology of creativity. Despite the relatively small size of the corpora used, the resulting set of 694 creativity words is sufficiently rich for this purpose, but is still somewhat large to work with in its raw form. In previous, related work \cite{jordanous10a} an attempt was made to identify key components by manually clustering creativity words by inspection of the raw data. In practice, this proved laborious and made it impossible systematically to consider all of the identified words. It also raised issues of subjectivity and experimenter bias. These problems are addressed here, at least in part, by automatically clustering the words according to a statistical measure of {\em distributional similarity\/} \cite{lin98}.  The more manageable collection of clusters may then be examined to identify key components or dimensions of creativity.

The intuition underlying distributional measures of similarity derives from the {\em distributional hypothesis\/} due to Harris \cite{harris68}. This hypothesis  states that similarity of distribution correlates with similarity of meaning: two words that tend to appear in similar linguistic contexts will tend to have similar meanings. The notion of linguistic context here is not fixed and might plausibly be modelled in a variety of different ways. For example, two words might be considered to inhabit the same context if they appear in the same document or the same sentence or if they stand in the same grammatical relationship to some other word (e.g. both occur as {\em subject\/} of a particular verb or {\em modifier\/} of a given noun). In practice it has been shown that modelling distribution in terms of grammatical relations leads to a tighter correlation between distributional similarity and closeness of meaning \cite{kilgarriff00}. 

In the present work, grammatical relations are used to represent linguistic context  and distributional similarity is measured as a function of the number of  relations 
that two words share. To illustrate, evidence that the words {\em concept} and {\em idea} are similar in meaning might be provided by occurrences such as the following: 

\begin{quote}
(1) the {\em concept/idea} involves \hfill (subject of verb `involve')\\
(2) applied the {\em concept/idea\/} \hfill (object of verb `apply') \\
(3) the basic {\em concept/idea\/} \hfill (modified by adjective `basic')
\end{quote}

\noindent
Grammatical relations were obtained from an analysis of the written portion of the British National Corpus \cite{leech92}, which had previously been processed using the RASP toolkit \cite{briscoe06} in order to extract them. Using this data, each word in the creativity corpus was associated with a list of all of the grammatical relations with which it occurred, together with their corresponding counts of occurrence.\footnote{Not all of the grammatical relation information output by RASP was used to calculate distributional similarity. In practice, just the subject, object and modifier relation types are used as these tend to give the best results \cite{weeds03}.} A potential difficulty with obtaining word similarity data based on the BNC (i.e. using data from sources of everyday usage of English, rather than from more specialist sources) would arise if the majority of the creativity words were used with distinctive or technical senses within the creativity corpus. From inspection and from knowledge of creativity literature, however, this situation was found to be unlikely. While some narrowly specialised usage may be present to some small degree in the set of creativity words, most words retain general senses as reflected in the wider BNC data set. An advantage of using the BNC is that its  size increases the chances of a comprehensive coverage of the general senses of each word of interest.

Distributional similarity of two words is measured in terms of the similarity of their associated lists of grammatical relations. A variety of different methods for calculating distributional similarity have been investigated in the literature, including standard techniques such as the cosine measure \cite[for example]{mannschu99}. The present work adopts an information-theoretic measure due to Lin \cite{lin98}, which has been widely used in language processing applications as a means of automatically discovering semantic relationships between words. In comparison to other similarity measures it has been shown to perform particularly well as a means of identifying near-synonyms \cite{weedsweir03,mccanavi09}.

Similarity scores were calculated between all pairs of creativity words of the same grammatical category. That is, scores were obtained separately for pairs of nouns, verbs, adjectives and adverbs. For a given set of words, word pair similarity data calculated in this way can be conveniently visualised as an edge-weighted graph, where nodes  correspond to words and edges are weighted by similarity scores (for any score $>$ 0), as in Fig. \ref{conceptIdea}.

\begin{figure}
\caption{{\bf Word pairwise similarity data visualised as an edge-weighted graph.} \\Nodes  correspond to words and edges are weighted by similarity scores (for any score $>$ 0). }
%% REMOVE IMAGES FROM SUBMITTED VERSION
%%\includegraphics[width=\textwidth]{images/Fig4}
\label{conceptIdea}
\end{figure}

Graphical representations of the similarity data like that shown in Fig. \ref{conceptIdea} provide a useful basis for analysing the creativity words and identifying recurring themes or components of creativity. Two complementary methods for identifying key components of the data were adopted:

\begin{description}

\item[Clustering:] The graph clustering software {\em Chinese Whispers} \cite{biemann06} was used to automatically identify word clusters (groups of closely interconnected words) in the dataset. This algorithm uses an iterative process to group together graph nodes that are located close to each other. By grouping words with similar meanings, the number of data items was effectively reduced and themes in the data could be recognised more readily from each distinct cluster. A sample of some of the resulting clusters can be seen in Fig. \ref{clustersZoom}. %To carry out Chinese Whispers, a preprocessing step was taken to categorise words according to their parts of speech, using the RASP software \cite{carrollRASP}.

\item[Inspection:] To focus on the words most closely related to creativity, the top twenty creativity words (i.e. the twenty  words with the highest log likelihood scores) were selected. Each word was then visualised as the root node of its own individual subgraph using the graph drawing software {\em GraphViz}\footnote{$http://www.graphviz.org/$, last accessed August 2016}. In order to reduce the amount of data to be examined, similarity scores were discarded if they fell below a threshold value (adjusted manually for each graph to highlight the most strongly connected words). This made the size and complexity of the graphs smaller and therefore easier to inspect and analyse visually. Fig. \ref{clusterToComponents} illustrates, in diagram form, the process of using manual inspection to identify components. 

\end{description}

\begin{figure}
\caption{{\bf Sample of clusters produced by the Chinese Whispers clustering step.}}
%% REMOVE IMAGES FROM SUBMITTED VERSION
%%\includegraphics[width=\textwidth]{images/Fig5}
\label{clustersZoom}
\end{figure}  

\begin{figure}
\caption{{\bf Illustration of the process of using manual inspection for further clustering.}}
%% REMOVE IMAGES FROM SUBMITTED VERSION
%%\includegraphics[width=\textwidth]{images/Fig6}
\label{clusterToComponents}
\end{figure}

As part of the manual inspection process, candidate components were further considered in terms of the {\em Four Ps} of creativity \cite{rhodes61,mooney63,mackinnon70,kaufman09} described earlier in this paper. This additional analysis provided a means of identifying alternative perspectives and revealing subtle (but still important) aspects of creativity.  For example, {\em novelty} is commonly associated with the results of creative behaviour (Product): how novel is the artefact or idea that has been produced? However, we could similarly recognise as creative an approach to a task  (Process) that does things in a novel and different way.  Also, if a product is new in a particular environment (Press), then it may well be regarded as creative to those in that environment. Viewing {\em novelty} from the perspectives of Product, Process and Press uncovers these subtle and interlinked distinctions.

% Results and Discussion can be combined.
\section*{Results and Discussion}
\label{results}

\subsection*{Components of creativity}\label{componentsResults}

From the analysis steps described in the previous section it was possible to extract a set of fourteen key components of creativity. These components are summarised in Fig. \ref{components} 
and are presented in more detail below. The components contribute collectively to the overall concept and may be regarded as providing an {\em ontology of creativity\/}.  It is important to note, however that the fourteen components do not constitute a set of necessary and sufficient conditions for creativity, in all its possible manifestations. There are two reasons for this. Firstly, some of the components we have identified appear to be logically inconsistent with others in the set. Consider for example the apparent need for autonomous, independent behaviour identified in {\em Independence and Freedom\/} and contrast this with the requirement for social interaction implied by {\em Social Interaction and Communication\/}. %\hidden{These factors may appear logically opposed.}
Secondly, of course, creativity %\hidden{does not lend itself to formal and precise definition and 
%I added this bit to clarify the point a bit [does not lend itself to formal and precise definition and]
also manifests itself in rather different ways across different domains \cite{plucker04} and components will vary in importance, according to the requirements of a particular domain. As an illustration of this second point, %Can we explicitly relate this illustration to one or more of the components that have been identified?
creative behaviour in mathematical reasoning has more focus on finding a correct solution to a problem than is the case for creative behaviour in, say,  musical improvisation \cite{colton08,jordanous12jims}. 

\begin{figure} 
%% REMOVE IMAGES FROM SUBMITTED VERSION
\caption{{\bf The fourteen key components of creativity identified through an analysis of the word clusters.}}
\label{components}
\end{figure}

The following set of fourteen components is therefore presented as a collection of dimensions -- attributes, abilities and behaviours, etc. -- which contribute to our understanding of creativity. The components should be treated as {\em building blocks} for creativity that may be arranged in different ways and with different emphases to suit different modelling purposes. The analysis of creativity in terms of the dimensions should be informative for a human audience and provide a basis for machine-understanding of the concept. %This approach is informed by debates in cognitive linguistics, computational linguistics and through philosophical underpinnings (see Background section). 
 Each component is presented here with a brief explanation or gloss. These explanations will later be used for part of the semantic content in the creativity ontology.

%The current work illuminates the sorts of issues that arise in formal modelling of subjective or `soft' concepts such as creativity, which are unlikely to resolve to fixed, static meaning(s). For example,  some components appear  logically inconsistent with others in the set: e.g. the need for autonomous, independent behaviour ({\em Independence and Freedom\/}) versus the requirement for social interaction ({\em Social Interaction and Communication\/}). Also, creativity clearly manifests itself in different ways across different domains \cite{plucker04} and components will vary in importance, according to the requirements of a particular domain. For example, creative behaviour in mathematical reasoning has more focus on finding a correct solution to a problem than is the case for creative behaviour in, say,  musical improvisation \cite{colton08}. The set of creativity components are therefore presented as a collection of dimensions -- attributes, abilities and behaviours, etc. -- which contribute to an overall improvement in the understanding of creativity that provides greater clarity and information both to a human audience and in machine-readable form for future Linked Data links and applications on the Semantic Web.  This approach is informed by debates in cognitive linguistics, computational linguistics and through philosophical underpinnings (section \ref{cogPhil}).

\begin{description}

\item[Active Involvement and Persistence:] \mbox{}\\
{\em Being actively involved; reacting to and having a deliberate effect on the creative process.\/}\\
{\em The tenacity to persist with the creative process throughout, even during problematic points\/}.

\item[Dealing with Uncertainty:] \mbox{}\\{\em
Coping with incomplete, missing, inconsistent, contradictory, ambiguous and/or uncertain information. Element of risk and chance - no guarantee that information problems will be resolved.\\
Not relying on every step of the process to be specified in detail; perhaps even avoiding routine or pre-existing methods and solutions.
}

\item[Domain Competence:] \mbox{}\\{\em
Domain-specific intelligence, knowledge, talent, skills, experience and expertise.\\
Knowing a domain well enough to be equipped to recognise gaps, needs or problems that need solving and to generate, validate, develop and promote new ideas in that domain.
}

\item[General Intellectual Ability:] \mbox{}\\{\em
General intelligence and IQ.\\
Good mental capacity.
}

\item[Generation of Results:] \mbox{}\\
{\em Working towards some end target, goal, or result.\/}\\
{\em Producing something (tangible or intangible) that previously did not exist.\/}

\item[Independence and Freedom:] \mbox{}\\{\em 
Working independently with autonomy over actions and decisions.\\
Freedom to work without being bound to pre-existing solutions, processes or biases; perhaps challenging cultural or domain norms.
}

\item[Intention and Emotional Involvement:] \mbox{}\\{\em
Personal and emotional investment, immersion, self-expression and involvement in the creative process.\\
The intention and desire to be creative: creativity is its own reward, a positive process giving fulfilment and enjoyment.
}

\item[Originality:] \mbox{}\\{\em
Novelty and originality; a new product, or doing something in a new way; seeing new links and relations between previously unassociated concepts.\\
Results that are unpredictable, unexpected, surprising, unusual, out of the ordinary.
}

\item[Progression and Development:] \mbox{}\\{\em
Movement, advancement, evolution and development during a process.\\
Whilst progress may or may not be linear, and an actual end goal may be only loosely specified (if at all), the entire process should represent some progress in a particular domain or task. }

\item[Social Interaction and Communication:] \mbox{}\\{\em
Communicating and promoting work to others in a persuasive and positive manner.\\
Mutual influence, feedback, sharing and collaboration between society and individual.
}

\item[Spontaneity/Subconscious Processing:] \mbox{}\\{\em 
No need to be in control of the whole process; thoughts and activities may inform the process subconsciously without being inaccessible for conscious analysis, or may receive less attention than others. \\
Being able to react quickly and spontaneously when appropriate, without needing to spend too much time thinking about the options.
}

\item[Thinking and Evaluation:] \mbox{}\\{\em 
Consciously evaluating several options to recognise potential value in each and identify the best option, using reasoning and good judgement.\\
Proactively selecting a decided choice from possible options, without allowing the process to stagnate under indecision.
}

\item[Value:] \mbox{}\\{\em 
Making a useful contribution that is valued by others and recognised as an achievement and influential advancement; perceived as special, `not just something anybody would have done'.\\
The end product is relevant and appropriate to the domain being worked in.
}

\item[Variety, Divergence and Experimentation:] \mbox{}\\{\em 
Generating a variety of different ideas to compare and choose from, with the flexibility to be open to several perspectives and to experiment and try different options out without bias.\\
Multi-tasking during the creative process.
}
\end{description}

\subsection*{Implementing a machine-readable ontology of creativity}\label{creativityOntology}

The fourteen components provide a fuller and clearer account of the constituent parts of the concept of creativity. An important aim of the current work is to make the components available as a resource for other researchers in computational creativity and to provide a basis for the automated evaluation of creative systems. As a step in this direction, the components have been expressed in an open, machine-readable form within the Semantic Web. In this way, the characterization of the components benefits from and is enriched by concepts that are already represented within the Semantic Web. %To achieve this, SKOS (Simple Knowledge Organisation System) \cite{skos} was used as a basis for constructing an OWL ontology representation of creativity and its components, 

In particular, the components are linked to the data in WordNet \cite{fellbaum98}, a large lexical database of English that has recently been made available as a Semantic Web ontology \footnote{$http://wordnet.rkbexplorer.com/$, last accessed August 2016}. In WordNet, words are grouped by sense and interlinked by lexical and conceptual relations. Note that, although the WordNet definition of the word such as `creativity' is brief (`the ability to create'), its utility lies in how it is linked to various concepts, such as its sense, hyponyms, type, `gloss' (brief definition) and other related concepts. Each creativity component relates to a cluster of keywords from the original set of 694 creativity words. Following Linked Data principles, each can therefore be linked across the Semantic Web to an appropriate set of concepts from WordNet. In this way,  associated semantic information is provided for each component. 
%The SKOS ontology incorporates three main classes:  {\em skos:Concept}  (anything we may want to record information about), {\em skos:ConceptScheme} (a set that collectively defines a skos:Concept) and  {\em skos:Collection} (a collection of semantically-related information). An instance of {\em skos:ConceptScheme\/} was created as {\em CreativityComponents}, to represent the set of components that defines the {\em skos:Concept\/} of {\em Creativity}. Each component is represented as an individual {\em skos:Concept\/}. 

The resulting encoding can be visualised as a graph, as shown in Fig. \ref{ontologyGraph}. 
The data has also been published under an Open Data Commons Public Domain Dedication and Licence (PDDL) \cite{miller08} at:
%as an OWL/XML file at:
\begin{quote}
\texttt{\scriptsize http://purl.org/creativity/ontology}
\end{quote}
%under a Open Data Commons Public Domain Dedication and Licence (PDDL) \cite{miller08} .
%The {\em skos:Concept\/}  
The concept labelled {\em Creativity} has the unique URI: 
\begin{quote}
\texttt{\scriptsize http://purl.org/creativity/ontology\#Creativity}
\end{quote}%
Any Linked Data that needs to refer to the concept can use this identifier. 

\begin{figure}
\caption{{\bf The ontology of creativity generated from this work's results, in graph form.}}
%% REMOVE IMAGES FROM SUBMITTED VERSION
%%\includegraphics[scale=0.5]{images/Fig8}
\label{ontologyGraph}
\end{figure}

\section*{Evaluation}

From a practical stand-point, the current work is part of an overarching project engaged with the question of the evaluation of creativity, particularly computational creativity \cite{jordanous12cc}. It is clear that a rigorous and comparative evaluation process needs clear standards to use as guidelines or benchmarks \cite{torrance88,kaufman09}.  

The components of creativity in this paper have been employed in two case studies to evaluate the creativity of computational systems \cite{jordanousphd,jordanous12jims,jordanous16aisb}. In these case studies, evaluation was carried out using the three step approach of the Standardised Procedure for Evaluating Creative Systems (SPECS) \cite{jordanous12cc}:

\begin{quote}
\begin{enumerate}
\item Identify a definition of creativity that your system should satisfy to be considered creative:
	\begin{enumerate}
	\item What does it mean to be creative in a general context, independent of any domain specifics?
	\item What aspects of creativity are particularly important in the domain your system works in (and what aspects of creativity are less important in that domain)?
	\end{enumerate}
\item Using Step 1, clearly state what standards you use to evaluate the creativity of your system.
\item Test your creative system against the standards stated in Step 2 and report the results.
\end{enumerate}    
\end{quote}

In both case studies, the components of creativity were chosen as the way of characterising creativity for step 1a of SPECS, and were weighted according to their importance and relevance for creativity in the creative domains under study for each case study (step 1b of SPECS). Each component was treated as one standard to be used to evaluate the creativity of the creative systems in the case studies (step 2 of SPECS). Each case study system was then tested against each component using feedback provided by judges (step 3 of SPECS), resulting in a detailed set of evaluative feedback on the creativity of each system in the case studies.

Case Study 1 \cite{jordanousphd,jordanous12cc} evaluated the creativity of three different computational musical improvisation systems \cite{jordanous12cc}. 
Case Study 2 used the components of creativity in an evaluation scenario where information and time was limited for evaluation, to simulate the forming of first impressions and snapshot judgements of the creativeness of a given computational creativity system \cite{jordanousphd,jordanous16aisb}. 

The resulting component-based evaluation yielded detailed information about creative strengths and weaknesses of the systems under investigation, highlighting those components where a system performs strongly. Crucially, the evaluation feedback also highlighted areas where a given system performed poorly. For example, in the musical improvisation study, Case Study 1, we found that, in general, creativity could be improved most by improving performance in {\em Social Interaction and Communication}, {\em Intention and Emotional Involvement} and {\em Domain Competence} (the three components found to be most important for creativity in musical improvisation). Similarly, it is useful to be able to quickly obtain formative feedback on strengths and weaknesses in time-limited scenarios such as that replicated in Case Study 2 during the development of creative systems (when ongoing evaluation of progress ideally needs to be both timely and time-efficient). Insight can then be obtained on where future development effort is best spent. 

The results described above were compared with those obtained from applying other evaluation models and with surveys of people's opinions, where people were asked how creative they thought each system was. There was general agreement between evaluation approaches on the most and least creative systems. The approaches differed in the formative feedback they provided, particularly for identifying strengths of the system at being creative, and weaknesses of the system to be improved. The model of creativity offered in this paper gave the most detailed feedback, but required most information to be collected. 

To support the usefulness of having the components as a tangible characterisation of creativity, an interesting finding was made as part of the first case study, in a separate evaluation carried out: asking for people's opinions on how creative the musical improvisation systems were. A striking observation was that a number of participants called for the word ``creativity'' to be defined before they felt comfortable with the task and confident in evaluating creativity in this setting, even though participants reported feeling generally positive or at least neutral towards the concept of computational creativity. This challenges the generally held view that people have a  common-sense working definition of creativity, at least in the context of judgement and evaluation. A representation of creativity is useful to:

\begin{enumerate}
\item establish what it means for something to be deemed creative; and
\item identify appropriate evaluation standards that replicate typical human opinion on how creative something is or in comparing two or more creative systems.
\end{enumerate}

\subsection*{Conclusions and directions for future work}
\label{conclusions}

This paper has described the methods used to identify a set of {\em components of creativity} using corpus-based, statistical language processing techniques. The motivation for the work is the need for a shared, comprehensive and multi-perspective model of creativity. Such a model should be of great value to researchers investigating the nature of creativity and in particular those concerned with the evaluation of creative practice. More broadly, the inter-disciplinary approach described here exemplifies a general approach to the investigation and representation of semantically fuzzy and essentially-contested concepts. For this reason, we expect that it will interest researchers investigating computational methods for analysing and representing other such concepts.

Rather than attempting to provide a unitary account of creativity, our approach extracts common, underlying themes that transcend discipline or domain bias. Our point of departure is the observation that the vocabulary used in discussions of the nature of creativity may be analysed in order to throw light on our understanding of the concept and its key attributes. Using techniques from corpus linguistics and natural language processing (as described in the Methods section), key components of creativity have been identified. The results of this novel, empirical analysis (presented in the Results section) inform the development of an {\em ontology of creativity\/} comprising a set of fourteen distinct components (see Fig. \ref{components}).
It is noted that each component makes a separate contribution to the overall meaning of the concept. At the same time, because creativity manifests itself in different ways across different domains \cite{plucker04}, the individual components vary in importance and influence according to the requirements of a given domain. The components can be therefore be usefully thought of as 'building blocks' for the concept in its different manifestations. Taken together, the components make creativity more tractable to study and to evaluate. 

The fourteen components provide a multi-perspective model of creativity that has been successfully applied in a comparative analysis and evaluation of computational creativity systems \cite{jordanous12cc,jordanousphd,jordanous16aisb} (see the Discussions section). The outcome of the evaluation process provides relatively fine-grained information about the creative strengths of a given system. This information in turn evidences ways in which a system could be considered creative. In addition, evaluation based on the components is able to highlight areas of weakness. These can be used to inform future work aimed at further developing a system's creative potential.

The components have been published in an open, machine-readable format, making them freely available to the research community. This has a number of implications. First, the set of components may be readily elaborated, extended or amended by other researchers investigating the concept of creativity. Second, the machine-readable format facilitates the development of creativity-aware applications, based on the components. Such applications might be developed to support manual evaluation of creative practice or as a significant step towards the development of methods for automated evaluation.

The problem of developing automated evaluation has elsewhere been described as `the Achilles' heel of AI research on creativity' \cite{boden99}.  An intriguing possibility that we are currently exploring is to further exploit language processing techniques to perform evaluation based on textual reviews, descriptions of system performance, or social media interactions \cite{jordanous15vem}. Such an approach would be analogous to the way sentiment analysis techniques are now in common use to evaluate attitude and opinion based on reviews of products or services \cite{panglee08}. This is a fascinating direction for future work, with great potential for real progress towards tackling computational creativity's `Achilles' heel'.

\section*{Supporting Information}

% Include only the SI item label in the subsection heading. Use the \nameref{label} command to cite SI items in the text.

\subsection*{S1 Appendix}\label{S1_Appendix}
{\bf Creativity Corpus.}  The following 30 papers were used as the {\em creativity corpus} for this work:

\begin{itemize}
\item T. M. Amabile. The social psychology of creativity: A componential conceptualization. Journal of Personality and Social Psychology, 45(2):357-376, 1983.
\item M. A. Boden. Precis of The Creative Mind: Myths and mechanisms. Behavioural and Brain Sciences, 17(3):519-570, 1994.
\item D. T. Campbell. Blind variation and selective retentions in creative thought as in other knowledge processes. Psychological Review, 67(7):380-400, 1960.
\item S. Colton, A. Pease, and G. Ritchie. The effect of input knowledge on creativity. In Proceedings of Workshop Program of ICCBR-Creative Systems: Approaches to Creativity in AI and Cognitive Science, 2001.
\item M. Csikszentmihalyi. Motivation and creativity: Toward a synthesis of structural and energistic approaches to cognition. New Ideas in Psychology, 6(2):159-176, 1988.
\item M. Dellas and E. L. Gaier. Identification of creativity: The individual. Psychological Bulletin, 73(1):55- 73, 1970.
\item A. Dietrich. The cognitive neuroscience of creativity. Psychonomic Bulletin \& Review, 11(6):1011-1026, 2004.
\item G. Domino. Identification of potentially creative persons from the adjective check list. Journal of Consulting and Clinical Psychology, 35(1):48-51, 1970.
\item W. Duch. Intuition, insight, imagination and creativity. IEEE Computational Intelligence Magazine, 2(3):40-52, 2007.
\item C. S. Findlay and C. J. Lumsden. The creative mind: Toward an evolutionary theory of discovery and innovation. Journal of Social and Biological Systems, 11(1):3-55, 1988.
\item C. M. Ford. A theory of individual creative action in multiple social domains. The Academy of Management Review, 21(4):1112-1142, 1996.
\item J. Gero. Creativity, emergence and evolution in design. Knowledge-Based Systems, 9(7):435-448, 1996. 
\item H. G. Gough. A creative personality scale for the adjective checklist. Journal of Personality and Social Psychology, 37(8):1398-1405, 1979.
\item J. P. Guilford. Creativity. American Psychologist, 5:444-454, 1950.
\item Z. Ivcevic. Creativity map: Toward the next generation of theories of creativity. Psychology of Aesthetics, Creativity, and the Arts, 3(1):17-21, 2009.
\item K. H. Kim. Can we trust creativity tests? A review of the Torrance tests of creative thinking (TTCT). Creativity Research Journal, 18(1):3-14, 2006.
\item L. A. King, L. McKee Walker, and S. J. Broyles. Creativity and the five-factor model. Journal of Research in Personality, 30(2):189-203, 1996.
\item R. R. McCrae. Creativity, divergent thinking, and openness to experience. Journal of Personality and Social Psychology, 52(6):1258-1265, 1987.
\item S. A. Mednick. The associative basis of the creative process. Psychological Review, 69(3):220-232, 1962. 
\item M. D. Mumford and S. B. Gustafson. Creativity syndrome: Integration, application, and innovation. Psychological Bulletin, 103(1):27-43, 1988. 
\item M. T. Pearce, D. Meredith, and G. A. Wiggins. Motivations and methodologies for automation of the compositional process. Musicae Scientae, 6(2):119-147, 2002.
\item J. A. Plucker, R. A. Beghetto, and G. T. Dow. Why isn't creativity more important to educational psychologists? Potentials, pitfalls, and future directions in creativity research. Educational Psychologist, 39(2):83-96, 2004.
\item R. Richards, D. K. Kinney, M. Benet, and A. P. C. Merzel. Assessing everyday creativity: Characteristics of the lifetime creativity scales and validation with three large samples. Journal of Personality and Social Psychology, 54(3):476-485, 1988.
\item G. Ritchie. The transformational creativity hypothesis. New Generation Computing, 24(3):241-266, 2006.
\item G. Ritchie. Some empirical criteria for attributing creativity to a computer program. Minds and Machines, 17:67-99, 2007.
\item D. L. Rubenson and M. A. Runco. The psychoeconomic approach to creativity. New Ideas in Psychology, 10(2):131-147, 1992.
\item M. A. Runco and I. Chand. Cognition and creativity. Educational Psychology Review, 7(3):243-267, 1995.
\item D. K. Simonton. Creativity: Cognitive, personal, developmental, and social aspects. American Psychologist, 55(1):151-158, 2000.
\item J. R. Suler. Primary process thinking and creativity. Psychological Bulletin, 88(1):144-165, 1980.
\item G. A. Wiggins. A preliminary framework for description, analysis and comparison of creative systems. Knowledge-Based Systems, 19(7):449-458, 2006.
\end{itemize}

\subsection*{S2 Appendix}\label{S2_Appendix}
{\bf Non-Creativity Corpus.} The following 60 papers were used as the {\em non-creativity corpus} for this work:

\begin{itemize}
\item C. Ames and J. Archer. Achievement goals in the classroom: Students' learning strategies and motivation processes. Journal of Educational Psychology, 80(3):260-267, 1988. % Cited by:  (since 1996) 639.
\item J. Anderson and D. Gerbing. Structural equation modeling in practice: A review and recommended two-step approach. Psychological Bulletin, 103(3):411-423, 1988. % Cited by:  (since 1996) 3884.
\item J. Arnett. Emerging adulthood: A theory of development from the late teens through the twenties. American Psychologist, 55(5):469-480, 2000. % Cited by:  (since 1996) 800.
\item M. Arulampalam, S. Maskell, N. Gordon, and T. Clapp. A tutorial on particle filters for online nonlinear/non-gaussian bayesian tracking. IEEE Transactions on Signal Processing, 50(2):174-188, 2002. % Cited by:  (since 1996) 1984.
\item A. Baddeley. Exploring the central executive. Quarterly Journal of Experimental Psychology Section A: Human Experimental Psychology, 49(1):5-28, 1996. % Cited by:  (since 1996) 672.
\item T. Baker, M. Piper, D. McCarthy, M. Majeskie, and M. Fiore. Addiction motivation reformulated: An affective processing model of negative reinforcement. Psychological Review, 111(1):33-51, 2004. % Cited by:  (since 1996) 177.
\item P. Barnett and I. Gotlib. Psychosocial functioning and depression: Distinguishing among antecedents, concomitants, and consequences. Psychological Bulletin, 104(1):97-126, 1988. % Cited by:  (since 1996) 366.
\item J. Baron. Nonconsequentialist decisions. Behavioral and Brain Sciences, 17(1):1-42, 1994. % Cited by:  (since 1996) 68.
\item F. Beach. The snark was a boojum. American Psychologist, 5(4):115-124, 1950. % Cited by:  (since 1996) 32.
\item M. Belkin, P. Niyogi, and V. Sindhwani. Manifold regularization: A geometric framework for learning from labeled and unlabeled examples. Journal of Machine Learning Research, 7:2399-2434, 2006. % Cited by:  (since 1996) 145.
\item G. Bonanno. Loss, trauma, and human resilience: Have we underestimated the human capacity to thrive after extremely aversive events? American Psychologist, 59(1):20-28, 2004. % Cited by:  (since 1996) 352.
\item T. Chan and L. Vese. Active contours without edges. IEEE Transactions on Image Processing, 10(2):266-277, 2001. % Cited by:  (since 1996) 1520.
\item H. Cheng and J. Schweitzer. Cultural values reflected in Chinese and U.S. television commercials. Journal of Advertising Research, 36(3):27-45, 1996. % Cited by:  (since 1996) 88.
\item C. Coello Coello. Evolutionary multi-objective optimization: A historical view of the field. IEEE Computational Intelligence Magazine, 1(1):28-36, 2006. % Cited by:  (since 1996) 109.
\item D. Comaniciu and P. Meer. Mean shift: A robust approach toward feature space analysis. IEEE Transactions on Pattern Analysis and Machine Intelligence, 24(5):603-619, 2002.  % Cited by:  (since 1996) 1613
\item L. Cronbach and L. Furby. How we should measure `change': Or should we? Psychological Bulletin, 74(1):68-80, 1970. % Cited by: (since 1996) 526.
\item M. Davis. Measuring individual differences in empathy: Evidence for a multidimensional approach. Journal of Personality and Social Psychology, 44(1):113-126, 1983. % Cited by:  (since 1996) 631.
\item J. Dem\~{s}ar. Statistical comparisons of classifiers over multiple data sets. Journal of Machine Learning Research, 7:1-30, 2006. % Cited by:  (since 1996) 389.
\item M. Dorigo, M. Birattari, and T. St\"{u}tzle. Ant colony optimization artificial ants as a computational intelligence technique. IEEE Computational Intelligence Magazine, 1(4):28-39, 2006. % Cited by:  (since 1996) 134.
\item E. Fischer and J. Turner. Orientations to seeking professional help: Development and research utility of an attitude scale. Journal of Consulting and Clinical Psychology, 35(1 PART 1):79-90, 1970. % Cited by:  (since 1996) 189.
\item J. Gibson. Observations on active touch. Psychological Review, 69(6):477-491, 1962. % Cited by:  (since 1996) 179.
\item A. Gopnik and J. Astington. Children's understanding of representational change and its relation to the understanding of false belief and the appearance-reality distinction. Child development, 59(1):26-37, 1988. % Cited by:  (since 1996) 314.
\item S. Gosling, S. Vazire, S. Srivastava, and O. John. Should we trust web-based studies? A comparative analysis of six preconceptions about internet questionnaires. American Psychologist, 59(2):93-104, 2004. % Cited by:  (since 1996) 313.
\item J. Gray. The psychophysiological basis of introversion-extraversion. Behaviour Research and Therapy, 8(3):249-266, 1970. % Cited by:  (since 1996) 225.
\item B. Grosz and S. Kraus. Collaborative plans for complex group action. Artificial Intelligence, 86(2):269- 357, 1996. % Cited by:  (since 1996) 275.
\item P. Groves and R. Thompson. Habituation: A dual-process theory. Psychological Review, 77(5):419-450, 1970. % Cited by:  (since 1996) 375.
\item M. Hall, J. Anderson, S. Amarasinghe, B. Murphy, S.-W. Liao, E. Bugnion, and M. Lam. Maximizing multiprocessor performance with the SUIF compiler. Computer, 29(12):84-89, 1996. % Cited by:  (since 1996) 161.
\item F. Happ\'{e}. The role of age and verbal ability in the theory of mind task performance of subjects with autism. Child development, 66(3):843-855, 1995. % Cited by:  (since 1996) 285.
\item C. Harland. Supply chain management: Relationships, chains and networks. British Journal of Management, 7(SPEC. ISS.):S63-S80, 1996. % Cited by:  (since 1996) 172.
\item S. Hayes, K. Strosahl, K. Wilson, R. Bissett, J. Pistorello, D. Toarmino, M. Polusny, T. Dykstra, S. Batten, J. Bergan, S. Stewart, M. Zvolensky, G. Eifert, F. Bond, J. Forsyth, M. Karekla, and S. McCurry. Measuring experiential avoidance: A preliminary test of a working model. Psychological Record, 54(4):553-578, 2004. % Cited by:  (since 1996) 168.
\item J. Hirsch and K. L\"{u}cke. Overview no. 76. mechanism of deformation and development of rolling textures in polycrystalline f.c.c. metals-i. description of rolling texture development in homogeneous cuzn alloys. Acta Metallurgica, 36(11):2863-2882, 1988. % Cited by:  (since 1996) 243.
\item P. Killeen. Mathematical principles of reinforcement. Behavioral and Brain Sciences, 17(1):105-172, 1994. % Cited by:  (since 1996) 86.
\item P. Kirschner, J. Sweller, and R. Clark. Why minimal guidance during instruction does not work: An analysis of the failure of constructivist, discovery, problem-based, experiential, and inquiry-based teaching. Educational Psychologist, 41(2):75-86, 2006. % Cited by:  (since 1996) 139.
\item S. Liao. Notes on the homotopy analysis method: Some definitions and theorems. Communications in Nonlinear Science and Numerical Simulation, 14(4):983-997, 2009. % Cited by:  (since 1996) 56.
\item C. Lord, L. Ross, and M. Lepper. Biased assimilation and attitude polarization: The effects of prior theories on subsequently considered evidence. Journal of Personality and Social Psychology, 37(11):2098- 2109, 1979. % Cited by:  (since 1996) 523.
\item S. Luthar, D. Cicchetti, and B. Becker. The construct of resilience: A critical evaluation and guidelines for future work. Child Development, 71(3):543-562, 2000. % Cited by:  (since 1996) 566.
\item G. Mandler. Recognizing: The judgment of previous occurrence. Psychological Review, 87(3):252-271, 1980. % Cited by:  (since 1996) 877.
\item G. Miller and J. Selfridge. Verbal context and the recall of meaningful material. The American journal of psychology, 63(2):176-185, 1950. % Cited by:  (since 1996) 57.
\item S. Miller. Monitoring and blunting: Validation of a questionnaire to assess styles of information seeking under threat. Journal of Personality and Social Psychology, 52(2):345-353, 1987. % Cited by:  (since 1996) 385.
\item G. Navarro and V. M\"{a}kinen. Compressed full-text indexes. ACM Computing Surveys, 39(1), 2007. % Cited by:  (since 1996) 62.
\item M. Nissen and P. Bullemer. Attentional requirements of learning: Evidence from performance measures. Cognitive Psychology, 19(1):1-32, 1987. % Cited by:  (since 1996) 729.
\item J. Payne, J. Bettman, and E. Johnson. Adaptive strategy selection in decision making. Journal of Experimental Psychology: Learning, Memory, and Cognition, 14(3):534-552, 1988. % Cited by:  (since 1996) 300.
\item J. Prochaska, C. DiClemente, and J. Norcross. In search of how people change: Applications to addictive behaviors. American Psychologist, 47(9):1102-1114, 1992. % Cited by:  (since 1996) 2742.
\item T. Richardson, M. Shokrollahi, and R. Urbanke. Design of capacity-approaching irregular low-density parity-check codes. IEEE Transactions on Information Theory, 47(2):619-637, 2001. % Cited by:  (since 1996) 1162.
\item W. Rozeboom. The fallacy of the null-hypothesis significance test. Psychological Bulletin, 57(5):416-428, 1960. % Cited by:  (since 1996) 133.
\item C. Rusbult. A longitudinal test of the investment model: The development (and deterioration) of satisfaction and commitment in heterosexual involvements. Journal of Personality and Social Psychology, 45(1):101-117, 1983. % Cited by:  (since 1996) 301.
\item T. Ryan. Significance tests for multiple comparison of proportion, variance, and other statistics. Psychological Bulletin, 57(4):318-328, 1960. % Cited by:  (since 1996) 154.
\item W. Schultz. Behavioral theories and the neurophysiology of reward. Annual Review of Psychology, 57:87-115, 2006. % Cited by:  (since 1996) 226.
\item E. Sirin, B. Parsia, B. Grau, A. Kalyanpur, and Y. Katz. Pellet: A practical OWL-DL reasoner. Web Semantics, 5(2):51-53, 2007. % Cited by:  (since 1996) 125.
\item T. Srull and R. Wyer. The role of category accessibility in the interpretation of information about persons: Some determinants and implications. Journal of Personality and Social Psychology, 37(10):1660- 1672, 1979. % Cited by:  (since 1996) 374.
\item J. Steiger. Tests for comparing elements of a correlation matrix. Psychological Bulletin, 87(2):245-251, 1980. % Cited by:  (since 1996) 705.
\item L. Steinberg, S. Lamborn, S. Dornbusch, and N. Darling. Impact of parenting practices on adolescent achievement: authoritative parenting, school involvement, and encouragement to succeed. Child development, 63(5):1266-1281, 1992. % Cited by:  (since 1996) 435.
\item D. Tao, X. Li, X. Wu, W. Hu, and S. Maybank. Supervised tensor learning. Knowledge and Information Systems, 13(1):1-42, 2007. % Cited by:  (since 1996) 54.
\item A. Tellegen, D. Lykken, T. Bouchard Jr., K. Wilcox, N. Segal, and S. Rich. Personality similarity in twins reared apart and together. Journal of Personality and Social Psychology, 54(6):1031-1039, 1988. % Cited by:  (since 1996) 398.
\item L. Thomas and D. Ganster. Impact of family-supportive work variables on work-family conflict and strain: A control perspective. Journal of Applied Psychology, 80(1):6-15, 1995. % Cited by:  (since 1996) 339.
\item I. Thompson. Coupled reaction channels calculations in nuclear physics. Computer Physics Reports, 7(4):167-212, 1988. % Cited by:  (since 1996) 360.
\item E. Tulving. Subjective organization in free recall of ``unrelated'' words. Psychological Review, 69(4):344- 354, 1962. % Cited by:  (since 1996) 121.
\item U. Von Luxburg. A tutorial on spectral clustering. Statistics and Computing, 17(4):395-416, 2007. % Cited by:  (since 1996) 107.
\item J. Williams, A. Mathews, and C. MacLeod. The emotional stroop task and psychopathology. Psychological Bulletin, 122(1):3-24, 1996. % Cited by:  (since 1996) 716.
\item J. Wright, A. Yang, A. Ganesh, S. Sastry, and Y. Ma. Robust face recognition via sparse representation. IEEE Transactions on Pattern Analysis and Machine Intelligence, 31(2):210-227, 2009. % Cited by:  (since 1996) 46.
\end{itemize}

\section*{Acknowledgments}
We would like to acknowledge Nick Collins and Chris Thornton for their helpful comments during this work. 
\pagebreak

\nolinenumbers

%\section*{References}
% Either type in your references using
% \begin{thebibliography}{}
% Text
% \end{thebibliography}
%
% OR
%
% Compile your BiBTeX database using our plos2015.bst
% style file and paste the contents of your .bbl file
% here.
% 

\bibliography{references}

%\begin{thebibliography}{10}
%\bibitem{bib1}
%Devaraju P, Gulati R, Antony PT, Mithun CB, Negi VS. Susceptibility to SLE in South Indian Tamils may be influenced by genetic selection pressure on TLR2 and TLR9 genes. Mol Immunol. 2014 Nov 22. pii: S0161-5890(14)00313-7. doi: 10.1016/j.molimm.2014.11.005
%
%\bibitem{bib2}
%Huynen MMTE, Martens P, Hilderlink HBM. The health impacts of globalisation: a conceptual framework. Global Health. 2005;1: 14. Available: http://www.globalizationandhealth.com/content/1/1/14.
%
%\end{thebibliography}

\end{document}